\documentclass{article}
\usepackage{spconf,amsmath,amssymb,graphicx,booktabs,multirow,xspace}
\usepackage{tikz}
\usepackage[T1]{fontenc}
\usetikzlibrary{arrows.meta,positioning,fit}

\newcommand{\ours}{MammoSeek\xspace}  % TODO: final system name

\title{MammoSeek: A Mammography Agent with selective zooming for grounded reports}
\name{Haoyan Ding$^{1,2}$, Qian Wang$^{1,2}$, Lixia Wang$^{2}$, Debiao Li$^{1,2}$ }
\address{$^{1}$University of California, Los Angeles, Los Angeles, CA, USA \\
$^{2}$Biomedical Imaging Research Institute, Cedars-Sinai Medical Center, Los Angeles, CA, USA \\}

\begin{document}
\maketitle

%=====================================================================
\begin{abstract}
Radiologists read a screening mammogram by first forming a global impression of the four views and then selectively magnifying suspicious regions before assigning a BI-RADS assessment. MLLMs, in contrast, see the whole image at a fixed, reduced resolution, which makes sub-millimeter calcifications invisible, and they perform close to chance on mammography. Therefore, we present \ours, a mammography agent that
reproduces this reading process. 
After forming an impression of each breast, the agent performs selective zooming: it decides by itself which detector candidates deserve a closer look and
inspects only those on full-resolution views. Guided by a BI-RADS assessment skill, it then assigns a category and generates a localized, lexicon-structured report for each breast; its assessment
is combined with image-model scores in a decision layer to decide recall.
On an internal (EMBED) and an external (VinDr-Mammo) dataset, \ours reaches exam-level AUCs of 0.824 and 0.818, far above other MLLMs, mammography agent and CNN classifiers. %, while producing structured reports with a descriptor-level F1 of 0.243/0.384. 
Selective zooming also outperforms both no zooming and forced zooming in localization, structured-report F1 and recall decisions at low cost.
\end{abstract}


%=====================================================================
\section{Introduction}
\label{sec:intro}
Screening mammography is read at high volume, and its output is not only a malignancy score: for each breast the
radiologist records a BI-RADS assessment that decides whether the woman is recalled, together with the findings that
motivate it, described in the BI-RADS lexicon and localized by clock position~\cite{birads}. An
assessment category is required in every mammography report~\cite{mqsa}, and recall decisions vary widely between
readers: only 59\% of radiologists reach the recommended abnormal interpretation rate~\cite{lehman2017}. Reaching
these decisions requires reading at several scales, because microcalcifications, an early sign of many cancers, are
often smaller than one millimeter. Radiologists therefore first survey the four views (craniocaudal, CC, and
mediolateral oblique, MLO, of each breast) to form an overall impression~\cite{kundel2007holistic,evans2013gist},
then magnify the regions that look suspicious~\cite{fallenberg2014mag}, confirm them on the other projection, and
finally assign a BI-RADS category following well-defined assessment conventions.

Deep learning classifiers trained on large screening cohorts predict malignancy with high
accuracy~\cite{nyubcc,gmic}, but they address only part of this task. They output a score rather
than a BI-RADS assessment with described, localized findings, so the report is still left to the radiologist.
Moreover, the score has to be turned into a recall decision by a threshold, and a threshold chosen at one site does
not transfer: at the 90\%-specificity threshold of our development data, the public NYU classifier~\cite{nyubcc}
keeps its AUC on an external dataset but recalls only 24\% of the suspicious breasts.

Multimodal large language models (MLLMs) can write such reports and explain them, but general and medical MLLMs
perform close to chance on mammography~\cite{mllmbench}, partly because the whole image is downsampled to the model's
input resolution, where calcifications disappear. Tool-using agents can give an MLLM access to specialized image
models, yet the mammography agent, MammoClaw~\cite{mammoclaw}, found that its tools did not reliably improve
performance, leaving open how an agent should use such tools. 

What is missing is a system that delivers both a reliable recall decision and a report grounded in what was actually
seen. We present \ours, in which an agent reads the exam the way a radiologist does---survey, selective
magnification, BI-RADS assessment---while the recall decision is taken by a decision layer calibrated on clinical
outcomes that combines the agent's assessment with image-model scores.
Our contributions are:
(i) a mammography agent that forms a global impression, decides by itself which candidates to magnify, confirms
them on high-resolution views and assigns BI-RADS assessments guided by a clinical assessment skill;
(ii) a magnified view designed for mammography that preserves the bright end of the intensity range so that
calcifications remain recognizable, together with lesion-level classifiers trained on detector-like boxes;
(iii) an evaluation on internal and external datasets covering diagnosis, BI-RADS agreement, localization and structured-report accuracy, showing that selective zooming outperforms both no and forced zooming, and that the agent is effective and robust
across datasets.

\begin{figure*}[t]
\centering
\includegraphics[width=1\textwidth]{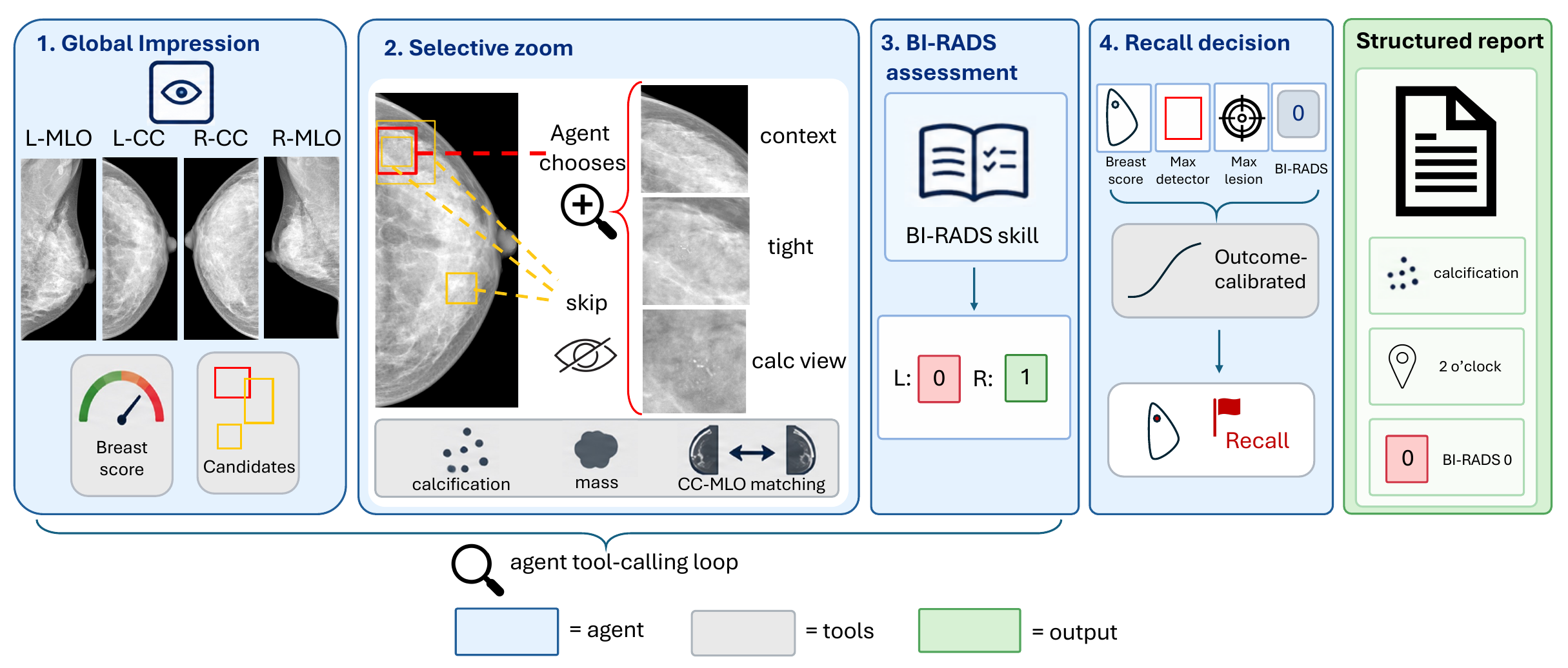}
\vspace{-24pt}
\caption{\ours follows the radiologist's reading process. The MLLM agent (blue) calls tools (grey) at each step.  Steps 1--3 form a tool-calling loop whose length and zoom targets are chosen by the agent. The recall decision combines the agent's BI-RADS assessment with
image-model scores; recalled breasts are reported with the agent's findings.}
\label{fig:overview}
\end{figure*}

%=====================================================================
\section{Method}
\label{sec:method}

\subsection{Overview}
\ours reads a four-view exam with a local MLLM (Qwen3-VL-8B~\cite{qwen3vl}) in a tool-calling loop (at most ten
rounds) that mirrors the clinical workflow (Fig.~\ref{fig:overview}): global impression (Sec.~\ref{sec:impr}),
selective zooming (Sec.~\ref{sec:zoom}), BI-RADS assessment (Sec.~\ref{sec:skill}), and recall decision with report
assembly (Sec.~\ref{sec:decision}). For each breast the agent returns a BI-RADS category, a list of findings (type,
BI-RADS descriptors, location, referenced candidate) and a recommendation.

\subsection{Global impression and candidate generation}
\label{sec:impr}
The agent first receives the four views in a canonical orientation and writes an impression for each breast (negative, benign finding, or suspicious finding with its approximate location). It can then query two whole-image
tools: a breast-level malignancy score from a linear probe on Mammo-CLIP~\cite{mammoclip} features, and a list of
candidate findings from a Faster R-CNN~\cite{fasterrcnn} detector trained on annotated EMBED regions, with position,
size in millimeters and detector score. As in a radiologist's reading, this survey only directs attention: most candidates are false positives, and the agent decides which ones deserve a closer look.

\subsection{Selective zooming}
\label{sec:zoom}
Zooming is an action of the agent. Given its impression and the candidate list, the agent calls \texttt{zoom} on a candidate of its choice, as many times as it deems necessary. The tool returns three images rendered from the full-resolution image: a tight crop, a context crop showing the surrounding tissue, and a
calcification view. Standard percentile windowing clips the bright end of the intensity range, which turns
clustered calcifications into a uniform white blob that resembles a mass; the calcification view keeps the full
bright range at near-native resolution (up to 1024\,px for a 5\,mm-margin crop), so that individual calcifications
and their morphology remain visible. The tool also returns the scores of two lesion-level heads: a malignancy
classifier and a calcification-vs-mass type head. Both are trained on randomly jittered boxes (scale
1--2.5$\times$, shift up to 30\%) and on detector false positives, so that training boxes match the detector boxes the
agent zooms into at test time. Finally, \texttt{cc\_mlo\_match} triangulates a finding on the other projection from its distance to the nipple and
gives its clock position and quadrant; the agent uses it to confirm a finding on both views.

\subsection{BI-RADS assessment skill}
\label{sec:skill}
Skills are short natural-language procedures that are inserted into the agent's context. Our BI-RADS assessment skill encodes
the conventions of the ACR BI-RADS Atlas~\cite{birads}: a screening exam is assessed with BI-RADS 0 (recall for additional imaging), 1 or 2
only; categories 3--5 are reserved for diagnostic work-up; each breast is assessed independently; and a finding that
cannot be dismissed as benign is assessed as BI-RADS 0. It also specifies the lexicon used for the findings
(e.g., calcification morphology and distribution, mass shape and margin).

\begin{figure*}[t]
\centering
\includegraphics[width=\textwidth]{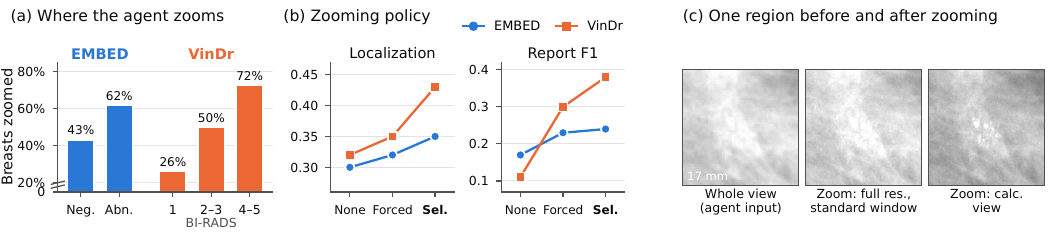}
\vspace{-28pt}
\caption{Selective zooming. (a) Fraction of breasts the agent zooms into (EMBED: negative vs. abnormal; VinDr: BI-RADS category).
(b) Localization and report F1 with no zooming, forced zooming and selective zooming decided by the agent. (c) A cluster of suspicious
calcifications as the agent sees it in the whole view, in a full-resolution crop with standard windowing%(the calcifications merge into a bright blob)
, and in the calcification view.}
\label{fig:zoom}
\end{figure*}

\subsection{Recall decision and report assembly}
\label{sec:decision}
MLLMs are poorly calibrated when asked for probabilities. We therefore take the recall decision with a logistic regression over four per-breast features: the breast-level score, the highest detector score, the highest lesion
malignancy score among candidates, and the agent's BI-RADS category. It is fitted on 1,812 EMBED development exams
with pathology-linked labels, using five-fold patient-grouped cross-fitting so that all tool outputs on the
development set are out-of-fold. The recall threshold is set at 90\% specificity on the development set. For a
recalled breast, the report contains the agent's findings, with the location of the main finding taken from
\texttt{cc\_mlo\_match}; if the agent reported no finding, the highest-scoring candidate is reported.

%=====================================================================
\section{Experiments}
\label{sec:exp}
\textbf{Data.} EMBED~\cite{embed} (Emory University, FFDM) was split by patient. The internal test set contains 444
four-view screening exams (111 malignant, 111 biopsy-benign, 222 negative). A breast is malignant if pathology on the
same side confirms cancer within 365 days; negative breasts are BI-RADS 1/2 with at least 300 days of
follow-up. The external test set is the official VinDr-Mammo~\cite{vindr} test split (375 exams), where a breast is positive if its consensus BI-RADS is 4 or 5. 

\textbf{Baselines.} Qwen2.5-VL-7B~\cite{qwen25vl}, Qwen3-VL-32B~\cite{qwen3vl} and the medical MLLM
Lingshu-7B~\cite{lingshu} answer directly from the four views with the same output schema; MammoClaw~\cite{mammoclaw}
is reimplemented from the paper with the same Qwen3-VL-8B backbone. As image classifiers that predict the outcome
directly, we run the public NYU BI-RADS classifier~\cite{nyubirads} and breast cancer classifier
(image+heatmaps)~\cite{nyubcc} with their released weights, and the Mammo-CLIP linear probe~\cite{mammoclip}.

\textbf{Metrics.} Exam-level AUC (malignant vs.\ negative exams; exam score is the maximum over both breasts).
At breast level: sensitivity and specificity of recall and their Youden index; quadratic-weighted
$\kappa$ between the predicted BI-RADS and the original radiologist assessment (EMBED) or dataset BI-RADS (VinDr);
localization (Loc: fraction of positive breasts that are recalled with the main finding's box or clock/quadrant
correct); and structured-report F1 (Rep: the main finding of a breast with a lesion is correct if its type matches the
annotation and, for calcifications, its morphology group (suspicious vs.\ typically benign) is correct; any finding
on a negative breast counts as a false positive). 

%=====================================================================
\section{Results}
\label{sec:results}

\begin{table*}[t]
\centering
\caption{Diagnosis, BI-RADS agreement, localization and structured reporting on EMBED (internal) and VinDr-Mammo
(external). AUC: exam level. Sens/Spec: breast-level recall. Rep: structured-report F1.}
\label{tab:main}
\setlength{\tabcolsep}{3.4pt}
\begin{tabular}{l|ccccc|ccccc}
\toprule
& \multicolumn{5}{c|}{EMBED (internal)} & \multicolumn{5}{c}{VinDr-Mammo (external)} \\
Method & AUC & Sens/Spec & $\kappa$ & Loc & Rep & AUC & Sens/Spec & $\kappa$ & Loc & Rep \\
\midrule
Qwen2.5-VL-7B~\cite{qwen25vl} & .535 & .06/.98 & .02 & .02 & .15 & .519 & .02/1.0 & .02 & .00 & .14 \\
Qwen3-VL-32B~\cite{qwen3vl} & .630 & .36/.89 & .20 & .18 & .12 & .609 & .14/.97 & .16 & .08 & .12 \\
Lingshu-7B~\cite{lingshu}$^{\ddagger}$ & .491 & 1.0/.00 & .00 & -- & .08 & .511 & 1.0/.00 & .00 & -- & .25 \\
NYU BI-RADS~\cite{nyubirads}$^{\S}$ & .615 & .00/1.0 & .09 & -- & -- & .606 & .00/1.0 & .01 & -- & -- \\
NYU cancer classifier~\cite{nyubcc} & .796 & .45/.90 & .33$^{*}$ & -- & -- & .797 & .24/.99 & .25$^{*}$ & -- & -- \\
Mammo-CLIP probe~\cite{mammoclip} & .790 & .45/.87 & .29$^{*}$ & -- & -- & .780 & .20/.99 & .23$^{*}$ & -- & -- \\
MammoClaw~\cite{mammoclaw}$^{\dagger}$ & .562 & .12/.93 & .03 & .00 & .00 & .531 & .09/.95 & .08 & .00 & .00 \\
\midrule
\ours (8B) & \textbf{.824} & .50/.90 & \textbf{.33} & \textbf{.35} & \textbf{.24} & \textbf{.818} & .51/.89 &
\textbf{.43} & \textbf{.43} & \textbf{.38} \\
\bottomrule
\end{tabular}
\\[2pt]{\small $^\dagger$Reimplementation; outputs no findings. $^\ddagger$Assesses every breast as BI-RADS 4
with a constant location (Loc not meaningful). $^{\S}$Predicts screening BI-RADS 0/1/2 but almost never BI-RADS 0.
$^{*}$Binary BI-RADS from a 90\%-specificity threshold on the development set.}
\end{table*}

\begin{table*}[t]
\centering
\caption{Ablations of the agent, the assessment skill and the zooming policy.  Zoom: number of zooms per exam. Values are EMBED / VinDr.}
\label{tab:abl}
\setlength{\tabcolsep}{4pt}
\begin{tabular}{l|cccccc}
\toprule
Configuration & AUC & Youden & $\kappa$ & Loc & Rep & Zoom \\
\midrule
w/o agent (decision layer) & .811 / \textbf{.828} & .379 / .240 & .32 / .25$^{*}$ & .35 / .25 & .22 / .12 & -- \\
w/o assessment skill & .805 / .820 & .397 / .352 & .30 / .39 & .34 / .34 & .24 / .29 & 1.9 / 1.4 \\
No zoom & .822 / .823 & .381 / .302 & .30 / .41 & .30 / .32 & .17 / .11 & -- \\
Forced zoom & .823 / .820 & .378 / .337 & .31 / .38 & .32 / .35 & .23 / .30 & 2.2 / 1.9 \\
\textbf{\ours (selective zoom)} & \textbf{.824 } / .818 & \textbf{.403 / .397} & \textbf{.33 / .43} & \textbf{.35 / .43} &
\textbf{.24 / .38} & \textbf{1.0 / 0.8} \\
\bottomrule
\end{tabular}
\\[2pt]{\small $^{*}$Binary BI-RADS (recall vs.\ negative).}
\end{table*}
% \vspace{-20pt}

\textbf{Comparison with existing methods.} 
As shown in Table~\ref{tab:main}, \ours achieves the highest exam-level AUC on both
datasets (0.824/0.818), above general and medical MLLMs, the existing mammography agent and image classifiers that
predict the outcome directly. General and medical MLLMs answering from the whole
images, as well as the reimplemented MammoClaw, are close to chance (AUC $\le$0.63): they recall very few breasts or
every breast and rarely localize a finding. Image classifiers that predict the outcome directly behave differently.
The NYU BI-RADS classifier, which predicts the screening assessment from the images, ranks exams only moderately
(AUC 0.62/0.61) and almost never assigns BI-RADS 0. The NYU cancer classifier and the Mammo-CLIP probe (which \ours
also uses as a tool) rank exams well (AUC 0.80/0.80 and 0.79/0.78), but output only a score, without BI-RADS
findings or locations, and their development-set threshold does not transfer: on VinDr they recall only 24\% and 20\% of BI-RADS 4--5 breasts at 99\% specificity. \ours not only reaches the highest AUCs, but also keeps a balanced operating point on both datasets (sensitivity 0.50/0.51 at specificity 0.90/0.89), agrees best with the reference BI-RADS, and localizes and
describes findings far more accurately than the others.

\textbf{Selective zooming} (Table~\ref{tab:abl}, Fig.~\ref{fig:zoom}). Zooming changes what the agent can see. In the downsampled
whole view, a cluster of calcifications spans only a few dozen pixels and is lost, and a full-resolution crop with
standard windowing merges it into a bright blob; only the calcification view resolves the individual calcifications
(Fig.~\ref{fig:zoom}c). Accordingly, without zooming the agent almost never recognizes calcifications, and localization and report accuracy are lowest
(Fig.~\ref{fig:zoom}b). 
More zooming is not necessarily better. Compared with selective zooming, forced zooming increases reading time by about 25\% while yielding lower localization, structured-report F1, Youden index, and $\kappa$ on both datasets.
When the agent decides, it zooms where it matters: its zoom rate rises with the suspicion of the breast.
With about one zoom per exam, selective zooming thus gives the best localization and reports (Loc 0.35/0.43, Rep 0.24/0.38).

\textbf{Role of the agent and the skill.} We next examine the contribution of the agent beyond the image models. Removing the agent has little effect on exam-level AUC but reduces structured-report F1, particularly on VinDr (0.12 vs. 0.38) and makes the recall decision overly conservative on external data, where the image-model scores are shifted (sensitivity/specificity 0.28/0.97 vs 0.51/0.89). 
Incorporating the agent's assessment improves the Youden index (EMBED $+0.024$, VinDr $+0.157$). Without the assessment skill the agent
zooms almost twice as often yet agrees less with the reference BI-RADS ($\kappa$ 0.30/0.39 vs.\ 0.33/0.43) and
localizes and describes external findings less accurately (Loc 0.34 vs.\ 0.43, Rep 0.29 vs.\ 0.38 on VinDr).
%=====================================================================
\section{Discussion and Conclusion}
\label{sec:conclusion}
Reading a mammogram like a radiologist---survey first, then magnify selectively, then assess by BI-RADS
conventions---turns an MLLM into a useful mammography agent. In \ours the image models set the diagnostic accuracy,
while the agent contributes what they cannot: localized, lexicon-structured findings and BI-RADS assessments that keep
the recall decision robust under domain shift. Selective zooming is essential to this role and is both more accurate
and cheaper than exhaustive inspection of candidates. Several limitations warrant consideration. VinDr-Mammo provides radiologist BI-RADS
rather than pathology, so part of the external gain may reflect closer agreement with radiologists; diagnostic
accuracy is bounded by the plug-in image models. Future work will evaluate the system on additional pathology-confirmed cohorts and conduct a radiologist reader study. We also plan to investigate stronger image models.

%=====================================================================
\section{Compliance with Ethical Standards}
\label{sec:ethics}
This retrospective study used de-identified data from the EMBED and VinDr-Mammo dataset, in accordance with their respective data use agreements. No additional ethical approval was required for this secondary analysis.

\section{Acknowledgments}
\label{sec:ack}
The authors declare no relevant conflicts of interest.

\bibliographystyle{IEEEbib}
\bibliography{refs}
\end{document}